\documentclass[11pt,letterpaper]{article}

\usepackage{cogsys}
\usepackage[T1]{fontenc}
\usepackage{times}
\usepackage{graphicx}
\usepackage{natbib}
\usepackage{amsmath}
\usepackage{booktabs}
\usepackage{array}
\usepackage{float}
\usepackage{xcolor}
\usepackage{tikz}
\usepackage{url}
\usetikzlibrary{arrows.meta,positioning,fit,calc}

\definecolor{mapgreen}{HTML}{009E73}
\definecolor{maporange}{HTML}{D97706}
\definecolor{mapblue}{HTML}{0072B2}
\definecolor{mapgray}{HTML}{5F6368}
\definecolor{paleblue}{HTML}{EAF4FA}
\definecolor{palegreen}{HTML}{EAF7F2}
\definecolor{paleorange}{HTML}{FFF3E3}
\definecolor{palegray}{HTML}{F1F3F4}

\tikzset{
  paperstage/.style={draw=mapgray, rounded corners=2pt, align=center,
    minimum width=1.62in, minimum height=0.63in, inner sep=5pt,
    font=\small},
  paperflow/.style={-{Latex[length=2.0mm,width=1.3mm]}, line width=0.8pt,
    draw=mapgray},
  papernode/.style={draw=mapgray, rounded corners=1.5pt, align=center,
    inner sep=3.5pt, font=\small, fill=white},
  exactlink/.style={-{Latex[length=1.8mm,width=1.2mm]}, draw=mapgreen,
    line width=1.1pt},
  structurallink/.style={-{Latex[length=1.8mm,width=1.2mm]}, draw=maporange,
    line width=1.1pt, dashed},
  feedback/.style={-{Latex[length=1.8mm,width=1.2mm]}, draw=mapblue,
    line width=0.9pt, dashed}
}

\ShortHeadings{Structure-Mapping-Guided Self-Explanation}{S. Lee, C. J. MacLellan, and D. Weitekamp}

\begin{document}

\title{Structure-Mapping-Guided Self-Explanation for Learning Mathematical Procedures}

\author{Shinhaeng Lee}{slee3487@gatech.edu}
\author{Christopher J. MacLellan}{cmaclell@gatech.edu}
\author{Daniel Weitekamp}{weitekamp@gatech.edu}
\address{Georgia Institute of Technology, Atlanta, GA 30332 USA}
\vskip 0.2in

\begin{abstract}
Worked examples are a powerful form of instruction, but learners must infer how
the demonstrated steps were produced.  A naive simulation of this
self-explanation process can generate thousands of numerical explanations that
reproduce one observed change without capturing its underlying procedure.  We
propose \emph{structure-mapping-guided self-explanation} as a computational
account of the cognitive biases that reduce search effort and make this
inference tractable.  The model represents mathematical expressions as typed
relational structures and uses structure mapping to identify corresponding
source and target regions.  For each changed target value, the corresponding
source region serves as an anchor: it guides abductive search toward
structurally relevant values and operations before broader alternatives,
yielding ordered, executable candidate procedures with inspectable source
evidence.  Across 70 mathematical transformations containing 120 changed numeric
components, the model recovered every intended procedure.  It returned the
intended procedure before any other computation producing the same target value
in 104 subproblems (86.7\%), compared with a median of 32 (26.7\%) across 100
unguided runs that tested candidate calculations in random order.  Our proposed
model also tested 93.8\% fewer combinations of values and operations than
unguided search before reaching the intended procedures.  These results provide
an efficient, interpretable account of how relational
structure can guide procedural learning and a testable hypothesis about human
self-explanation from worked examples.
\end{abstract}

\section{Introduction}

Worked examples let students encounter a procedure before they can state or
execute it independently.  Learning from such an example requires more than
recording its final answer.  A learner must determine which parts of the
initial expression correspond to parts of the result, explain the changes, and
construct candidate knowledge that can later be tested and applied.  Studies
of worked-example learning associate successful learning with
self-explanation, in which learners generate inferences that connect presented
steps to underlying principles
\citep{chi1989self,renkl1997worked}.  This poses a concrete computational
question: how do learners abduce useful explanations rather than the many
coincidental explanations that merely reproduce individual demonstrations?

Consider a demonstrated step from a Pythagorean calculation,
\begin{equation}
  c^2=5^2+12^2 \quad\longrightarrow\quad c^2=25+144.
  \label{eq:geometry-demo}
\end{equation}
The value 25 has more than one numerical explanation: the learner could evaluate
the displayed power $5^2$, or multiply the displayed 5 by itself.  Likewise,
144 can be obtained as either $12^2$ or $12\times12$.  Many less plausible
explanations become available if every number in the expression, along with
numbers and operations not shown there, is treated as equally relevant.  The
learner could combine the 5 in $5^2$ with the unrelated exponent 2 in $c^2$, or
produce 25 as $(-5)^2$ using a number absent from the demonstration.  These
programs return the correct values, but they do not preserve the operations and
the way their operands are used in the corresponding source regions.

More importantly, without a way to identify relevant operands and operations,
the learner must evaluate many possible combinations.  We compare our proposed
model with unguided search, which tests the same candidate calculations in
random order rather than using structural relationships to prioritize relevant
numbers and operations.  In the finite search
evaluated for this case, structure-mapping-guided search tested exactly two
arithmetic operations across the two changed values: one evaluation of each
corresponding power.  Unguided random ordering required a median of 884.  A
learner therefore needs a way to use relational structure before considering
the broader space of numerical explanations.

Treating known numbers and operations as an unstructured inventory is both
expensive and an unlikely account of how people approach a worked step: visible
operators, nesting, and mathematical roles cue what changed and which quantities
are relevant.  We therefore hypothesize that structure mapping between initial
and resulting expressions generates correspondences that order evidence during
self-explanation.  Search begins near a corresponding source region, moves
outward through the expression, and introduces absent knowledge only when
source-visible evidence is insufficient.

We operationalize this \emph{structure-mapping-guided self-explanation} account
with typed abstract syntax trees (ASTs), whose nodes expose operators, nesting,
and ordered argument roles: the positions or functions of components within an
operation, such as base versus exponent or first versus second addend.  Changed
target numbers become numerical self-explanation subproblems; search expands by
AST distance from their corresponding source regions; and successful
explanations are composed into executable candidate transformations.  Candidates
are retained only if they reconstruct the demonstrated target.  The resulting
ordered candidate space may isolate one procedure immediately or retain
alternatives for later evidence to distinguish.

Our goal here is not to outperform data-driven mathematical solvers, including
large language models (LLMs) developed for quantitative reasoning
\citep{lewkowycz2022solving}.  Rather, we seek a precise account of how novice
learners could narrow interpretations of worked examples without becoming
mired in thousands of admissible distractors.  This contrast also concerns
learning transparency: our model exposes source--target correspondences,
ordering biases, candidate alternatives, and executable symbolic procedures.
Although we focus on abduction from single examples, we conclude by situating
the mechanism within theories of inductive learning over multiple examples,
including Decomposed Inductive Procedure Learning (DIPL)
\citep{weitekamp2025dipl}.

Our primary evaluation asks whether the complete structure-mapping-guided
mechanism recovers demonstrated procedures earlier and with less search than an
unguided alternative.  Randomized and component-level controls then separate
the contributions of visible-first staging, mapping-relative distance, and
anchoring.  Finally, we test whether the local explanations compose into
complete procedures and whether those procedures apply to new values when their
source roles are supplied.  Across 70 transformations, the complete
structure-mapping-guided mechanism recovered all 120 intended local procedures,
returned the intended procedure before any alternative explanation of the
target in 86.7\% of subproblems, and examined 93.8\% fewer combinations of
values and operations.  Unguided search returned the intended procedure first
in a median of 26.7\% of subproblems across 100 random search orders.  Additional
analyses showed that this prioritization
persisted when local explanations were composed into complete procedures.

\section{Background and Related Work}

\subsection{Worked Examples and Self-Explanation}

Worked examples are commonly used alongside instruction and practice to show
how a procedure applies to a particular problem.  Their instructional benefit
depends on what a learner does with the presented steps.  \citet{chi1989self} found that more
successful learners generated explanations that connected example steps to
domain principles, and subsequent work characterized substantial individual
differences in this activity \citep{renkl1997worked}.  These findings motivate
self-explanation as a learning process, but they do not by themselves specify
an algorithm that chooses among candidate explanations.

The present work focuses on that search-control problem.  A numeric target can
have many numerically correct arithmetic derivations.  Counting every derivation
that returns the target as successful would conflate procedural learning with
coincidence.  Our model therefore requires explanations to draw on the
demonstrated source and evaluates them against intended procedures that specify
both an operation tree and exact source occurrences.  Structure does not
eliminate abduction; it constrains the order in which alternatives are
considered.

\subsection{Simulated Learners and Procedure Induction}

Computational accounts of learning procedures from instruction have a long
history.  SIERRA induced arithmetic procedures from examples while seeking to
model regularities and misconceptions in children's learning
\citep{vanlehn1990mindbugs}.  SimStudent learned cognitive skills from
demonstrated problem-solving steps \citep{matsuda2007simstudent},
and work on simulated-learner evaluation emphasized that models should be
assessed as learners rather than only as final performers
\citep{koedinger2015evaluating}.  The Apprentice Learner architecture separates
learning the computation for an action from learning where it applies and when
it should be selected \citep{maclellan2016apprentice}.  Decomposed Inductive
Procedure Learning (DIPL) extends this idea by coordinating specialized
inductive mechanisms to acquire educational procedures from small numbers of
examples \citep{weitekamp2025dipl}.

Using the how/where/when decomposition adopted by Apprentice Learner and DIPL
\citep{maclellan2016apprentice,weitekamp2025dipl}, our account focuses on
how-learning: inferring a computation that can produce the observed result.
Where-learning identifies the locations of the computation's arguments, while
when-learning determines the conditions under which the resulting rule should
be applied.  Unlike a solver supplied with the correct production rules, this
learner does not possess the demonstrated behavior before observing it; unlike
a system that stores only an input--output association, it constructs
inspectable symbolic candidates.  Section~6 returns to the broader architecture
that can store, test, apply, and refine them.

\subsection{Structure Mapping and Analogy}

Structure-mapping theory treats analogy as alignment of relational systems;
isolated attributes are secondary \citep{gentner1983structure}.
The Structure-Mapping Engine (SME) operationalizes this account
\citep{falkenhainer1989sme}.  It constructs and evaluates mutually consistent
correspondences.  Subsequent work has extended SME for large-scale cognitive
modeling \citep{forbus2017sme}.  Analogical systems have also used worked
solutions to formulate models for transfer in AP Physics
\citep{klenk2009physics}.

Our use of structure mapping differs from direct analogical transfer.  The
source and target here are two states within one demonstrated step.  Mapping
identifies which occurrences play corresponding roles and which target values
changed.  The correspondences then control a separate symbolic search for the
latent computation.  Thus structure mapping supplies a relevance heuristic
for self-explanation rather than a complete solution by itself.

\subsection{From Observations to Executable Knowledge}

ACT-R distinguishes declarative representations from procedural knowledge and
models proceduralization through mechanisms such as production compilation
\citep{anderson2007mind}.  Our model also produces executable knowledge
candidates, but it begins from an instance of a procedure being applied rather
than a complete declarative description of that procedure.  It must infer the
missing computation before a candidate transformation can be formed.

More broadly, computational models of scientific discovery have shown how
heuristic search can generate and test hypotheses about observed regularities
\citep{langley1987scientific}.  The mathematical setting here is
narrower, but the methodological commitment is similar: specify the
representation, operators, search control, and acceptance tests that turn
observations into explanatory knowledge.

\section{Structure-Mapping-Guided Self-Explanation}

\begin{figure}[!t]
\centering
\begin{tikzpicture}[x=0.49in,y=0.36in]
  \node[font=\small\bfseries] at (0,0.75) {Source structure};
  \node[font=\small] at (0,0.18) {$c^2=5^2+12^2$};
  \node[font=\small\bfseries] at (8,0.75) {Target structure};
  \node[font=\small] at (8,0.18) {$c^2=25+144$};

  \node[papernode] (seq) at (0,-0.55) {$=$};
  \node[papernode] (sc) at (-0.85,-1.45) {$c^2$};
  \node[papernode] (splus) at (0.75,-2.15) {$+$};
  \node[papernode, fill=paleorange] (sfive) at (0.20,-2.90) {$5^2$};
  \node[papernode, fill=paleorange] (stwelve) at (1.30,-3.65) {$12^2$};

  \node[papernode] (teq) at (8,-0.55) {$=$};
  \node[papernode] (tc) at (7.15,-1.45) {$c^2$};
  \node[papernode] (tplus) at (8.75,-2.15) {$+$};
  \node[papernode, fill=paleorange] (ttwentyfive) at (8.20,-2.90) {$25$};
  \node[papernode, fill=paleorange] (tonefortyfour) at (9.30,-3.65) {$144$};

  \draw[draw=mapgray!65, line width=0.7pt] (seq) -- (sc);
  \draw[draw=mapgray!65, line width=0.7pt] (seq) -- (splus);
  \draw[draw=mapgray!65, line width=0.7pt] (splus) -- (sfive);
  \draw[draw=mapgray!65, line width=0.7pt] (splus) -- (stwelve);
  \draw[draw=mapgray!65, line width=0.7pt] (teq) -- (tc);
  \draw[draw=mapgray!65, line width=0.7pt] (teq) -- (tplus);
  \draw[draw=mapgray!65, line width=0.7pt] (tplus) -- (ttwentyfive);
  \draw[draw=mapgray!65, line width=0.7pt] (tplus) -- (tonefortyfour);

  \draw[exactlink] (seq.east) -- (teq.west);
  \draw[exactlink] (sc.east) -- (tc.west);
  \draw[exactlink] (splus.east) -- (tplus.west);
  \draw[structurallink] (sfive.east) -- (ttwentyfive.west);
  \draw[structurallink] (stwelve.east) -- (tonefortyfour.west);

  \node[papernode, fill=palegray, text width=2.65in, font=\scriptsize]
    at (1.35,-5.55)
    {\textbf{Unguided search}\\
     no mapping-based priorities\\
     consider all available number--operation\\
     combinations together};
  \node[papernode, fill=palegreen, text width=2.65in, font=\scriptsize]
    at (7.95,-5.55)
    {\textbf{Structure-mapping-guided search}\\
     $5^2\leftrightarrow25$: prioritize evaluating $5^2$\\
     $12^2\leftrightarrow144$: prioritize evaluating $12^2$\\
     consider corresponding operands and operations first};
\end{tikzpicture}
\caption{Structure mapping guides numerical self-explanation in the running
example.  Green arrows mark exact correspondences that are copied unchanged,
while dashed orange arrows mark changed structural correspondences.  These
mappings anchor 25 to $5^2$ and 144 to $12^2$, causing search to consider the
operands and operations in each corresponding source region before unrelated
alternatives.}
\label{fig:mechanism-example}
\end{figure}
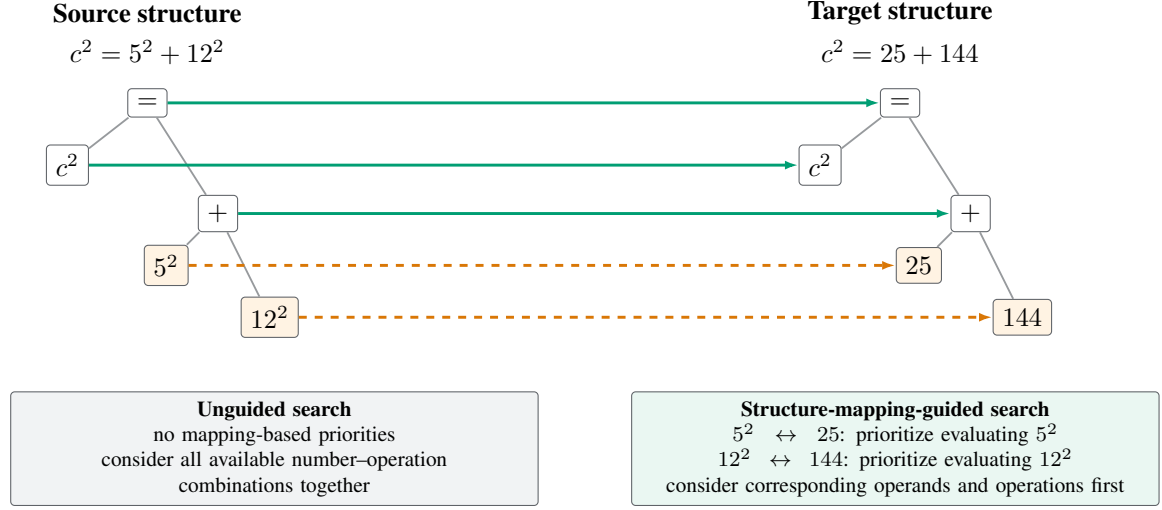

We introduce the mechanism through Equation~\ref{eq:geometry-demo}.  If every
source number, arithmetic operation, and known constant is immediately
available, numerical search can explain 25 and 144 in many coincidental ways.
Structure mapping instead identifies which source regions play the same roles
as those changed target values.  Figure~\ref{fig:mechanism-example} shows how
those correspondences constrain what search considers first.

The mechanism proceeds in four stages.  First, it represents the source and
target expressions relationally.  Second, it maps their corresponding
components.  Third, those correspondences order the operands and operations
considered during numerical self-explanation.  Finally, the local explanations
are composed into executable candidate procedures.  Following the theory--model
distinction of \citet{langley2018theories}, we specify this sequence as an
executable, testable cognitive hypothesis.

\subsection{Relational Representation and Structure Mapping}
\label{sec:mapping}

A raw string exposes characters, tokens, and written order, but not directly
the relational organization required to determine that a number is a
coefficient, exponent, numerator, or child of a particular operation.  The
model therefore parses each expression into a typed AST.  Node types include
numbers, symbols, addition, subtraction, multiplication, division, fractions,
and powers; ordered child positions make these roles explicit.  This is not a
complete representation of mathematical semantics, but it makes semantically
meaningful structure explicit enough for mapping and candidate ordering.

Each node occupies a relational position in the tree.  Identical numeric
values remain distinct when they occur in different mathematical roles.  For
example, an exponent of 2 and a denominator of 2 are separate occurrences
because they participate in different relations.  This distinction matters
because using the correct value from the wrong role can reproduce a target
number while supporting a different procedure.

We use Structure-Mapping with Tight Bounds (SMTB) to establish source--target
correspondences.  SMTB is the structure-mapping approach provided by the
Cognitive Rule Engine (CRE) \citep{weitekamp2026smtb}.  We use its default search settings.
SMTB identifies AST components compatible in type and relational role and
evaluates correspondences by agreement between the nodes and preservation of
surrounding relational structure.  Matching constants or symbols strengthen a
correspondence, but equality is not required: a compound source region can
correspond to a target value when both occupy corresponding roles.

We relax the usual one-to-one correspondence requirement of structural
consistency \citep{falkenhainer1989sme}.  Each source occurrence may correspond
to at most one target occurrence or remain unmatched, while several source
occurrences may correspond to the same target component.  The selected mapping
favors jointly preserved parent--child relations and ordered argument roles.

Our model then selects a primary source correspondence for each matched target
component.  Selection proceeds from parents to children: an established parent
correspondence restricts the available source region and identifies the expected
argument position.  For a changed numeric target, the model favors source
regions containing numerical evidence in that position.  Remaining ambiguity
is resolved by structural proximity, preference for source occurrences not
already selected, exact equality, and SMTB's original ordering.  Neither
intended-procedure annotations nor numerical explanation results influence
these decisions.  For a changed numeric target, the selected source region
becomes its initial anchor; Section~\ref{sec:anchors} describes anchoring and
structural distance.

In Figure~\ref{fig:mechanism-example}, corresponding addition nodes propose
$5^2\leftrightarrow25$ between their first arguments.  A power node and numeric
leaf have different concrete types, but both are expression nodes in the same
argument role under corresponding additions.  Relational context therefore
supports the correspondence without requiring numerical equality; the second
arguments similarly yield $12^2\leftrightarrow144$.

An \emph{exact correspondence} preserves structure and content and therefore
needs no numerical explanation.  In Figure~\ref{fig:mechanism-example}, the
equation relation, $c^2$, and addition are exact and are copied into the
candidate procedure.  A \emph{changed structural correspondence} links regions
that occupy corresponding roles but differ in content.  Thus
$5^2\leftrightarrow25$ and $12^2\leftrightarrow144$ identify the two numerical
self-explanation subproblems.  This relational criterion is more general than
matching raw tree positions: inserting, removing, or simplifying a surrounding
operation can change a component's literal path while preserving its role.
Structure mapping therefore identifies which target changes require
explanation and where the search for supporting evidence should begin.

\subsection{Anchors and Structural Distance}
\label{sec:anchors}

Each changed target value's corresponding source region becomes its initial
\emph{anchor}.  If that region is a number, the number itself is the anchor.  If
it is compound, the entire subtree is anchored, including its operation and
numeric descendants.  Thus, the anchor region $5^2$ supplies the exponentiation
relation and its operands, 5 and 2.  Such an anchor is \emph{executable} when
its inputs are numerical and its operations belong to the learner's arithmetic
knowledge.  The learner carries out the displayed calculation and compares its
result with the target: $5^2$ evaluates to 25, so it is retained as a candidate
explanation.  An expression such as $x^2$ cannot be evaluated numerically without
a value for $x$; its numerical descendants can still supply evidence for search.
By contrast, in $3x+5x\rightarrow8x$, structure
mapping identifies both source coefficients as possible correspondences for the
changed coefficient 8.  The model selects 3 as its initial numeric anchor based
on structural position, then considers the nearby coefficient 5 and addition,
yielding $3+5=8$.  Anchoring expresses the hypothesis that a changed value was
produced from the source region in the corresponding structural role.

Anchoring is an initial search hypothesis, not an inviolable constraint.  The
displayed calculation is tested first when a compound anchor is executable.
If it does not produce the target, the anchor's operands and operation
initialize a growing evidence pool.  A compound region that cannot be evaluated
directly supplies its available numerical and operational evidence to this
pool.  Additional source-visible values and operations enter in
structural-distance order, and each expansion enables new candidate
combinations without repeating combinations already tested.  For a numeric
anchor, the corresponding source number remains fixed while nearby
source-visible values and operations are admitted in the same order.  Knowledge
not displayed in the expression is considered only after visible evidence has
been exhausted; if those candidates also fail, another source number becomes
the fixed anchor and the process repeats.  This progressive relaxation prevents
a poor initial correspondence from blocking recovery while exhausting the
strongest structurally supported hypotheses first.

The order in which source values and operations enter during this relaxation is
determined by their structural distance from the anchor.  For AST nodes $u$ and
$v$, this distance is the number of edges from $u$ up to their lowest common
ancestor and then down to $v$:
\begin{equation}
 d(u,v) =
 \bigl(\operatorname{depth}(u)-\operatorname{depth}(\operatorname{lca}(u,v))\bigr)
 +\bigl(\operatorname{depth}(v)-\operatorname{depth}(\operatorname{lca}(u,v))\bigr).
 \label{eq:distance}
\end{equation}
For the anchor region $5^2$, exponentiation belongs to the region and its
operands 5 and 2 are one edge from the power node.  Reaching 12 requires moving
up to addition and down into the sibling power expression, so it enters later.
The measure provides one graded notion of relational locality rather than
separate rules for a parent, sibling, or whole expression.  Literal AST distance
is an operational approximation of the cognitive claim that explanation search
is sensitive to relational proximity.

\subsection{Prior Knowledge and Progressive Abductive Search}
\label{sec:search}

These anchor-relative priorities determine when evidence becomes available to
abductive search.  Occurrences of numbers and operations in the source enter the
candidate pool in increasing AST distance from the anchor, while equal-valued
occurrences in different structural roles remain distinct.  In the running
example, this ordering makes evaluation of the displayed $5^2$ the first candidate
explanation for 25.

The learner's arithmetic background knowledge consists of addition,
subtraction, multiplication, division, and exponentiation, together with the
integers from $-10$ through 10.  Operations or constants absent from the source
are available only during fallback.  The search is \emph{bounded} by this
finite vocabulary.  To keep the experimental hypothesis space controlled, each
candidate applies one arithmetic operation to two inputs.
The target value cannot be introduced as an unexplained constant, and every
candidate must use at least one occurrence from the source.

The abductive search constructs and tests candidate procedures one at a time.
Each candidate is executed, and candidates that produce the changed target
value are retained as explanations.  Candidate identity includes both the
operation and the exact source occurrences used as inputs.  Consequently,
$5^2$ and $5\times5$ remain different explanations of 25, as do explanations
that use equal-valued numbers from different structural positions.  Numerically
equal outputs do not cause one derivation to replace another.

Structure-mapping-guided search first tests the exact computation displayed by
an executable compound anchor.  It then expands the available values and
operations by structural distance and tests each newly enabled one-operation
candidate.  With a numeric anchor, the corresponding source number remains
fixed throughout its visible and fallback stages before another source
occurrence becomes the anchor.  A candidate enters search only after both
inputs and its operation are available, so its stage in the search order is
determined by its most distant required element.  Only candidates with
identical operations and source bindings are treated as duplicates.  No
domain-specific rule such as ``multiply a coefficient by an exponent'' is
encoded; source structure instead determines which general arithmetic
candidates are tested first.

\subsection{From Local Explanations to Procedures}

Exact correspondences are represented as copied components.  Each changed
numeric component is replaced by a variable in the target template, and each
symbolic explanation becomes a function over the source occurrences that it
used.  Filling the template with copied components and explanatory functions
composes local accounts into executable candidate transformations.

In the running example, the exact components define the target template
$c^2=H_1+H_2$.  The first hole, $H_1$, is filled by applying exponentiation to
the base and exponent from $5^2$; the second hole, $H_2$, is filled in the same
way from $12^2$.
The resulting executable transformation retains the copied equation and
addition structure while computing both changed values from their corresponding
source roles.  Executing a candidate on the demonstration and confirming that
it reconstructs the target is an internal consistency check, not a separate
search mechanism.  When several candidates reconstruct the target, they remain
provisional alternatives for later evidence to distinguish.

\subsection{Extending the Search Beyond Local Evidence}

Longer expressions offer more competing evidence even when the changed value
requires only one arithmetic operation.  Consider this evaluated example:
\begin{equation}
  3x+12x+\frac{18}{3}-2^4
  \quad\longrightarrow\quad
  15x+\frac{18}{3}-2^4.
  \label{eq:operator-example}
\end{equation}
The fraction and power are unchanged, while the two like terms become $15x$.
For target 15, the selected source anchor is the region $3x$.  Because $x$ has
no numerical value, this region cannot be evaluated directly.  It instead
supplies its coefficient 3 and multiplication as initial evidence.  Expanding
through the surrounding addition reaches coefficient 12 and supports the
explanation $3+12=15$.  The denominator 3 is a different occurrence in a more
distant region; using it merely because its value matches the coefficient
would obscure the demonstrated operand role.  Starting from the anchored source
region $3x$, our proposed model expands the candidate pool by structural
distance to include the nearby coefficient $12$.  It reaches $3+12=15$ after
testing five candidate calculations, compared with a median of 479 under
unguided search.  The preceding four calculations do not produce 15, making
$3+12$ the first successful explanation.  This illustrates how anchoring and
progressive expansion reduce search effort without requiring the learner to
know the correct computation in advance.  Section~\ref{sec:component-results} uses this family of examples to
examine the contribution of operation distance when several operation types
are visible.

The same search can recruit evidence absent from the source.  In the
demonstrated derivative step
\begin{equation}
  \frac{d}{dx}\,4x^6 \quad\longrightarrow\quad 24x^5,
  \label{eq:derivative-demo}
\end{equation}
structure mapping relates coefficient $4$ to 24, exponent 6 to 5, and $x$ to itself.  The
anchor strategy makes the search interpretable:
\begin{equation}
  \underbrace{4}_{\text{anchor}}
    \xrightarrow[\text{source exponent }6]{\times}24,
  \qquad
  \underbrace{6}_{\text{anchor}}
    \xrightarrow[\text{background constant }1]{-}5.
  \label{eq:anchor-derivative}
\end{equation}
The first explanation expands from the mapped coefficient to a more distant
source value and visible multiplication; the second recruits subtraction and 1
only after source-visible evidence is insufficient.  Together they support the
candidate procedure
\begin{equation}
  a x^b \quad\longrightarrow\quad (a b)x^{b-1}.
  \label{eq:generalized-derivative}
\end{equation}
No derivative rule was supplied in advance.  The model explains the observed
step using the same mapping, anchoring, and search processes as in the geometric
example, showing that the mechanism can recover coefficient and exponent
relations embedded in derivative notation.  It also does so efficiently: for
this case, our proposed model examined 62 value--operation combinations across the
two changed values, compared with a median of 327.5 under unguided random
ordering, an 81.1\% reduction.

\section{Evaluation Method}

\subsection{Transformation Corpora}

We evaluated the model on 70 source--target transformations assembled in two
corpora (Table~\ref{tab:corpus}).  The primary corpus contains 47 transformations
spanning four mathematical domains and three levels of structural distance.  An
extended corpus adds 23 more structurally demanding transformations adapted
from worked-example families in open educational mathematics texts.  To keep
the hypothesis space controlled and the comparisons interpretable, each changed
numeric component in the evaluated transformations has an intended explanation
that uses one arithmetic operation.  Cases vary local arithmetic reduction,
combination of values from different source regions, nesting, repeated values,
and the need to introduce an operation or constant not displayed in the
source.  Adaptations preserved the mathematical operation and operand roles of
the original step while expressing it in the model's relational vocabulary.
Each case represents one transition between adjacent worked steps rather than
an entire problem reduced directly to its final answer.  Each includes at least
one changed target number whose procedure can be identified from the
demonstrated step.  For example, $(9-3)/(5-2)$ transforms to $6/3$, preserving
the outer division while reducing its children.  This format tests explanation
of a visible change while retaining the surrounding relational context.  Local
search and complete-procedure composition were evaluated independently on all
70 cases.  A separate transfer test evaluated application to new numerical
instances when source roles were supplied.

\begin{table}[H]
\centering
\caption{Composition of the primary and extended corpora, with illustrative
adjacent transformations.}
\label{tab:corpus}
\small
\begin{tabular}{p{1.40in}rrp{2.65in}}
\toprule
Mathematical context & Primary & Extended & Example transformation \\
\midrule
Algebra/precalculus & 12 & 6 & $4(2x+3)\rightarrow 8x+12$ \\
Geometry & 12 & 6 & $(10-6)^2+(8-5)^2\rightarrow 4^2+3^2$ \\
Calculus notation & 11 & 5 & $\frac{d}{dx}4x^6\rightarrow24x^5$ \\
Vector/matrix notation & 12 & 6 &
$\langle2,3\rangle\!\cdot\!\langle5,4\rangle:
2\cdot5+3\cdot4\rightarrow10+12$ \\
\bottomrule
\end{tabular}
\end{table}

The operational representation covers arithmetic, equations, powers,
fractions, symbols, and parentheses.  Derivative, integral, vector, and matrix
contexts test scalar transformations embedded in richer mathematical notation;
for example, the final row of Table~\ref{tab:corpus} explains the component
products of a dot product.  The primary corpus contains 79 changed target
numbers and the extended corpus contains 41, for 120 numeric self-explanation
subproblems in total.  The primary corpus contains 16 local, 15 intermediate,
and 16 distant transformations; the extended corpus contains eight local, seven
intermediate, and eight distant transformations.  Both corpora vary simultaneous
changes, nesting, repeated values, introduced primitives, and the distance
between a mapped region and relevant source evidence.

\subsection{Intended Procedures}

Each changed target number defines one \emph{numeric self-explanation
subproblem}.  Before running the reported experiments, we specified its
intended procedure from the operation and operand roles expressed by the worked
step.  The specification records the arithmetic operation tree, the exact
source-number occurrences used as arguments, and any operation or constant
that the step requires but the source does not display.  For target 25 in
Equation~\ref{eq:geometry-demo}, the intended procedure is
to apply exponentiation to the base and exponent occurrences displayed in
$5^2$.
The numerically equivalent $5\times5$ returns 25, but it does not preserve the
displayed power and operand roles and therefore counts as a competing
explanation.

Candidate matching preserves source-occurrence identity: two written 2s in
different roles are not interchangeable merely because they have the same
value.  Argument order is canonicalized only for commutative addition and
multiplication, so $a+b$ and $b+a$ represent the same arithmetic procedure.
The AST otherwise remains as written; no general algebraic reduction makes
$x-1$ equivalent to $x+(-1)$.  These criteria let the experiment ask whether
search reaches the particular procedure and source roles expressed by the
worked step, rather than whether it can produce the same target number by any
means.

\subsection{Controlled Search Conditions}
\label{sec:conditions}

We evaluated six configurations of the same symbolic search process
(Table~\ref{tab:conditions}).  The primary comparison tests our complete
structure-mapping-guided model, labeled \emph{full guidance}, against unguided
search, which assigns no structural priorities.  A randomized visible-first control and three
component ablations assess mapping-relative order, anchoring, value order, and
operation order.  For each numeric
self-explanation subproblem, the source and target were parsed and
structure-mapped once;
configurations without mapping-based guidance ignored the resulting
correspondences.  All configurations used the same source occurrences,
operations, background integers from $-10$ to 10, grounding rule, and
one-operation candidate space.  Equal outputs never merged candidates with
different operations or source bindings.  The configurations differed only in
the order in which they tested those candidates.  Eventual coverage was
therefore identical by design.  Every candidate had to use at least one source
occurrence, so no configuration could explain a target solely by asserting a
background constant.

\begin{table}[H]
\centering
\caption{Controlled candidate-ordering configurations.  ``Distance'' orders
source occurrences by AST distance from the mapped region.  Random ordering
respects the priorities described below; it does not change the candidate
space.  Anchor-based conditions retain the initial preference for the displayed
calculation when the compound anchor is executable.}
\label{tab:conditions}
\small
\begin{tabular}{lccc}
\toprule
Configuration & Mapped anchor & Value order & Operation order \\
\midrule
Unguided search & No & Random & Random \\
Randomized visible-first & No & Visible first & Visible first \\
Anchor only & Yes & Random & Random \\
Value-distance guidance & Yes & Distance & Random \\
Operation-distance guidance & Yes & Random & Distance \\
Full guidance & Yes & Distance & Distance \\
\bottomrule
\end{tabular}
\end{table}

The six conditions differ as follows:
\begin{itemize}
  \item \textbf{Unguided search} tests candidates in random order without
  prioritizing a mapped anchor, nearby evidence, or knowledge displayed in the
  source.
  \item \textbf{Randomized visible-first} prioritizes candidates using displayed
  numbers and operations before those requiring absent knowledge.  It randomizes
  candidates within those priorities without using an anchor or AST distance.
  This control tests the simpler policy of trying displayed knowledge first.
  \item \textbf{Anchor only} prioritizes candidates supported by the initial
  mapped anchor, then uses random ordering within the resulting priority groups.
  It does not order additional evidence by structural distance.
  \item \textbf{Value-distance guidance} retains anchoring and orders candidates
  by the structural proximity of their source numbers.  Operations receive no
  additional distance-based priority.
  \item \textbf{Operation-distance guidance} retains anchoring and orders candidates
  by the structural proximity of their source operations.  Numbers receive no
  additional distance-based priority.
  \item \textbf{Full guidance} combines anchoring with structural-distance
  ordering of both numbers and operations, following the progressive search in
  Sections~\ref{sec:anchors}--\ref{sec:search}.  For a numeric anchor, it exhausts
  the candidates using that number, including fallback knowledge, before
  moving to the next source number.
\end{itemize}

The four anchor-based conditions first test the displayed computation when
their compound anchor is executable.  Subsequent ordering follows the
priorities above.  Full guidance is deterministic; for each other condition,
we evaluated 100 independently seeded random orders consistent with its
constraints.  The intended procedure served only as the stopping criterion for
reported rank and search effort and did not influence candidate ordering.

The original transformations offered limited competition among visible
operation types.  We therefore conducted two targeted analyses of operation
guidance.  The first asked whether prioritization remained reliable with
\emph{broader background knowledge}: we expanded the operation vocabulary from
five to eight and reran all 70 cases with other controls fixed.  The second
asked whether operation distance helped when \emph{visible operations competed}.
It used 24 transformations, six per mathematical context, containing 27 numeric
subproblems.  Every source displayed at least four operation types, with the
intended operation in the corresponding structural region, as in
Equation~\ref{eq:operator-example}.  Expanding the background adds possible
operations; the second analysis additionally supplies structural evidence about
which visible operation is relevant.

\subsection{Complete Procedures and Transfer to New Values}

We tested whether local explanations could compose and transfer.  A complete
procedure selects one candidate for each changed target value; in the running
example, it pairs an explanation for 25 with one for 144.  Successful local
explanations retained the order in which each search configuration encountered them.  We
ordered complete procedures by the sum of their local rank positions.  When two
combinations had the same total, we preferred the combination that retained the
higher-ranked explanation for the earlier changed component in the target
expression.  Forty-one of the 70 transformations required more than one local
explanation.

Transfer used 16 families, four per mathematical context, each containing one
demonstration and two new instances with values in the same structural roles.
We applied the demonstration's top-ranked procedure unchanged to both.  We then
used the first new instance as additional evidence, selecting the earliest
demonstration candidate that also solved it and applying that candidate unchanged
to the second.  Holding input roles constant isolated procedure reuse and the
ability of another example to resolve ambiguity.

\subsection{Measures and Analysis}
\label{sec:measures}

A candidate matched the intended procedure only when its normalized operation
tree and exact source-occurrence bindings matched the specification
described above.  Returning the correct number was insufficient.  For instance,
$2/2$ can generate 1, but it does not match an intended procedure that obtains 1
by subtracting source occurrences 3 and 2.

We measured search effort as the number of \emph{value--operation combinations
examined} before reaching the intended procedure.  This count includes every
grounded candidate tested, including candidates that do not produce the target.
Because each candidate applies one operation, it is also the number of
arithmetic operations evaluated by the search.

Capability and selectivity were measured by intended-procedure coverage,
explanation rank, recall within the first five explanations, and the number of
alternative explanations before the intended procedure.  An alternative
explanation is a different program that also produces the target value.
\emph{Intended first}, or rank 1, means that the intended procedure is the first
successful explanation, even if unsuccessful calculations were tested earlier.
For example, a hypothetical search for 25 might test $5+2$, then $5\times5$,
then the intended $5^2$.  It has examined three calculations and found two
explanations; the intended one has rank 2, with one alternative before it.

For each search order, we total the number of subproblems with the intended
procedure found, first, or within the first five explanations.  We also sum
alternative explanations and examined combinations across all 120 subproblems.
For randomized conditions, tables report the median of each whole-corpus total
across 100 orders, rather than a median per subproblem.  Fractional medians can
therefore occur even though every individual run has integer counts.

For complete procedures, we report these measures over best-first
combinations.  Transfer is scored by exact reconstruction of a new target, so an
algebraically equivalent procedure is accepted even if its operation tree
differs from the intended one.

We report reductions in aggregate effort as
$1-E_{\mathrm{full}}/E_{\mathrm{comparison}}$.  The primary contrast between
full guidance and unguided search tests the complete structure-mapping-guided
policy, which coordinates anchoring, distance staging, and fallback.  The
supporting contrast with randomized visible-first search isolates
mapping-relative ordering beyond the general policy of considering displayed
knowledge first.  To estimate uncertainty, we draw 2,000 bootstrap resamples of
the 70 whole transformations, keeping the numeric subproblems from each
transformation together.  Each resample also selects one of the 100 random
orders for the comparison condition.  The middle 95\% of the resulting
improvements defines the reported confidence interval.  Grouping by
transformation accounts for the related evidence shared by its subproblems.

For every subproblem, we recorded whether its anchor contained a source
occurrence used by the intended procedure.  This measure tests whether mapping
supplied relevant evidence without assuming one uniquely correct region
boundary.  We also recorded whether an executable compound anchor was evaluated
intact before any alternative candidate.

\section{Results}

\subsection{Finding Intended Explanations with Less Search}

Our proposed model found the intended procedures earlier while testing fewer
calculations.  All six configurations eventually recovered the same 120
intended procedures; the difference was how much search they required to reach
them (Table~\ref{tab:all-results}).  In the tables, \emph{full guidance} denotes
our proposed model.  Results for randomized conditions are medians across 100
search orders.

The model returned the intended explanation first for 104 of the 120 numeric
changes (86.7\%), compared with 32 under unguided search (26.7\%; range
25--39).  Here, \emph{first} means the first successful explanation of the
target, not necessarily the first calculation tested.  The difference was
60.0 percentage points (95\% confidence interval [CI]: 48.4--70.5).
The model also placed 118 intended procedures among the first five successful
explanations, compared with 82 under unguided search.

\begin{table}[H]
\centering
\caption{Results for 120 numeric changes.  ``Found'' counts recovered intended
procedures.  ``Intended first'' and ``Within 5'' count changes for which the
intended procedure appears first or among the first five successful
explanations.  Alternative explanations also produce the target but differ
from the intended procedure.  The last two columns total alternatives before
the intended procedure and all calculations tested through its recovery,
including unsuccessful attempts.  Both totals cover all 120 changes.
Randomized entries are
medians of whole-corpus totals over 100 orders; full guidance is deterministic.}
\label{tab:all-results}
\small
\setlength{\tabcolsep}{4pt}
\begin{tabular}{@{}lrrrrr@{}}
\toprule
Configuration & Found & \shortstack{Intended\\first} & \shortstack{Within\\5} &
\shortstack{Alternative explanations\\before intended} &
\shortstack{Calculations\\tested} \\
\midrule
Unguided search & 120 & 32 & 82 & 575.5 & 39,194.0 \\
Randomized visible-first & 120 & 62 & 107 & 230.5 & 6,713.5 \\
Anchor only & 120 & 98 & 118 & 44.0 & 3,897.0 \\
Value-distance guidance & 120 & 104 & 120 & 23.5 & 853.5 \\
Operation-distance guidance & 120 & 103 & 119 & 25.0 & 2,417.5 \\
Full guidance & 120 & 104 & 118 & 34.0 & 2,442.0 \\
\bottomrule
\end{tabular}
\end{table}

Reaching these explanations required substantially less computation.  Across
all 120 changes, the model tested 2,442 candidate calculations, compared with
39,194 under unguided search: a 93.8\% reduction (95\% CI: 89.8--96.7\%).
It also encountered fewer competing explanations that produced the target but
did not match the intended procedure: 34 rather than 575.5, a 94.1\% reduction.
The savings extended across individual transformations.  The median
per-transformation reduction in calculations was 99.6\%, with the middle half
of reductions ranging from 97.2\% to 99.7\%.

The model also improved on the simpler strategy of trying displayed knowledge
first.  Randomized visible-first search uses this strategy but does not order
evidence by its relationship to the changed target.  It returned the intended
explanation first in 62 subproblems (51.7\%; range 55--72), compared with the
model's 104.  The model exceeded every one of this control's 100 random orders;
its advantage over the median was 35.0 percentage points (95\% CI:
23.1--46.6).  It also tested 63.6\% fewer calculations (95\% CI: 47.6--75.4\%)
and encountered 85.2\% fewer competing explanations.  Thus, using the
source--target correspondence to anchor and order evidence added a benefit
beyond simply preferring knowledge visible in the source.  All four confidence
intervals favor the proposed model after accounting for variation across
transformations and comparison search orders.

\subsection{Combining and Reusing Learned Procedures}

The advantage extended from explaining individual numbers to explaining whole
transformations.  After combining the local explanations, both the proposed
model and unguided search recovered all 70 intended complete procedures.
The model ranked the intended complete procedure first in 56 transformations
(80.0\%), compared with an unguided median of 15.5 (22.1\%; range 11--19).
Among the 41 transformations with multiple changed values, the model ranked
30 intended complete procedures first, compared with a median of one under
unguided search.  Across all 70 transformations, it encountered 133 alternative
complete procedures before the intended ones, compared with 81,116 under
unguided search.  Prioritizing the local explanations therefore also reduced
how many unintended combinations were encountered when constructing a whole
procedure.

The learned procedures could also be reused with new numerical values.  In the
16 transfer families, the model's first candidate produced the correct target
for 28 of the 32 new instances, with corresponding input roles supplied.
It solved the second new instance in 14 of the 16 families without using the
first new instance to select a procedure.

One additional example resolved the remaining ambiguity in this transfer
corpus.  For each family, we used the first new instance to select the earliest
candidate that also solved it, then applied that candidate unchanged to the
second instance.  This succeeded in all 16 families.  The selected candidates
already appeared near the beginning of the original list: their median rank
was 1 and their maximum rank was 3.  Thus, the additional example distinguished
among existing candidates rather than requiring a new explanation search.
These tests establish procedure reuse and selection with input roles supplied;
learning where to find those inputs and when to apply a procedure requires the
broader architecture discussed in Section~6.2.

\subsection{Where the Search Savings Come From}
\label{sec:component-results}

Prioritizing source numbers by structural distance provided the largest search
savings on the original corpus.  Value-distance guidance retains the mapped
anchor but does not give operations an additional distance-based priority.
It matched the proposed model's 104 intended-first results while testing a
median of 853.5 calculations rather than 2,442.  It also placed all 120
intended procedures within the first five explanations, compared with 118 for
the proposed model.  It was therefore the least expensive configuration on
these problems.  Because it still uses structure mapping and AST distance,
this result locates the strongest benefit in the structural ordering of numbers.

Anchoring alone already returned a median of 98 intended procedures first,
using 3,897 calculations.  Operation-distance guidance returned a median of
103 first using 2,417.5 calculations.  These conditions also improved on
unguided search, but neither matched the search savings of value-distance
guidance (Table~\ref{tab:all-results}).

Mapping supplied relevant starting evidence: every initial anchor contained
an operand used by the intended procedure.  In 79 subproblems, the anchor was
an executable expression tested intact first; in 31, it was a number held
fixed during initial search.  The remaining 10 anchors were non-executable
regions that supplied numbers for search.  Nearby evidence more often led
directly to the intended explanation: it ranked first in 63 of 70 subproblems
whose farthest required operand was within one AST edge, 33 of 39 at two or
three edges, and 8 of 11 at four or more.  More distant explanations remained
reachable; distance changed their priority rather than excluding them.

Two additional analyses show when operation ordering contributes beyond value
ordering (Table~\ref{tab:operation-analyses}).  The first increases the number
of known operations on the same problems.  The second tests expressions with
several visible operations, so their structural positions can indicate which
one is relevant.

\begin{table}[H]
\centering
\caption{When operation-distance guidance helps.  The first two rows use the
same 70 transformations and 120 numeric changes; the third uses 24
transformations with 27 numeric changes and competing visible operations.
``Intended first'' counts changes for which the intended explanation appears
before any alternative.  Calculations tested are totals through recovery of
the intended procedures.  Full guidance is the proposed model; value-distance
entries are medians over 100 orders.}
\label{tab:operation-analyses}
\small
\setlength{\tabcolsep}{4pt}
\begin{tabular}{@{}>{\raggedright\arraybackslash}p{1.58in}rrrr@{}}
\toprule
& \multicolumn{2}{c}{Intended first} & \multicolumn{2}{c}{Calculations tested} \\
\cmidrule(lr){2-3}\cmidrule(l){4-5}
Analysis & Full guidance & Value-distance & Full guidance & Value-distance \\
\midrule
Original, 5 operations & 104/120 & 104/120 & 2,442.0 & 853.5 \\
Original, 8 operations & 104/120 & 101/120 & 3,889.0 & 1,317.5 \\
Competing visible operations & 27/27 & 26/27 & 149.0 & 263.5 \\
\bottomrule
\end{tabular}
\end{table}

With a larger operation vocabulary, the proposed model maintained its number
of intended-first results but required more calculations than value-distance
guidance.  Expanding the background from five to eight operations increased
the candidate space by 61.7\%.  The model still returned 104 intended
explanations first, while the value-distance median fell from 104 to 101.
However, it tested 3,889 calculations, compared with 1,317.5 for value-distance
guidance.  More available operations therefore produced a modest selection
benefit, not a search-cost advantage.

Absent operations help explain this tradeoff.  In 23 of the 120 subproblems,
the intended calculation required an operation not displayed in the source.
Such an operation has no position in the source AST, so structural distance
cannot prioritize it.  The proposed model tries visible operations before
introducing it, whereas value-distance guidance can try it earlier.  For
example, the subtraction needed to lower the exponent in
Equation~\ref{eq:derivative-demo} is absent from the source.  Preferring visible
operations adds work when the required operation is not among them.

When several visible operations competed, operation-distance guidance improved
both selection and search effort.  On the separate set of 27 numeric changes,
the proposed model returned all 27 intended explanations first and tested
149 calculations.  Value-distance guidance returned a median of 26 first and
tested 263.5 calculations.  The proposed model therefore used 43.5\% fewer
calculations.  For example, it reached $3+12$ in
Equation~\ref{eq:operator-example} after five calculations, compared with a
value-distance median of ten.

Together, these analyses distinguish two contributions: value distance
supplied most of the search savings on the original corpus, while operation
distance helped when the source provided structural evidence for choosing
among competing operations.  Simply adding more known operations did not make
the proposed model cheaper.  The comparison of five and eight operations does
not establish the tradeoff for still larger operation vocabularies.

\section{Discussion}

\subsection{A Computational Hypothesis About Learning}

The primary comparison supports the paper's central claim.  The complete
mechanism recovered all 120 intended local procedures, ranked 104 first rather
than the unguided median of 32, encountered 94.1\% fewer competing explanations,
and examined 93.8\% fewer candidate combinations.  Compared with randomized
visible-first search, it ranked 104 rather than a median of 62 first while
examining 63.6\% fewer combinations.  This stronger comparison shows that the
advantage comes not only from considering displayed knowledge first, but also
from ordering that knowledge according to its mapped structural relationship to
the changed target.  The ablations further indicate that value-distance
guidance supplies most of the search reduction on the original corpus and
matches full guidance's intended-first count at lower cost.  Operation
distance provides an additional benefit when competing visible operations offer
structural evidence about which calculation is relevant.  Together, these findings
show that the mechanism keeps intended procedures reachable while moving
structurally supported explanations toward the beginning of search.

These benefits were not limited to explaining changed values independently.
After the local explanations were composed, the intended complete procedure
ranked first in 56 of 70 transformations, compared with an unguided median of
15.5, and the mechanism encountered 99.8\% fewer preceding complete
alternatives.  Thus, the local priorities remained effective when several
explanations had to be combined into one procedure.  The composed procedures
could also be reused with new values: the top-ranked candidates produced 28 of
32 new targets.  When one additional instance was used to select among the
candidates generated from the original demonstration, the selected procedure
solved the second instance in all 16 families, with input roles supplied.  These
results show that the mechanism produces manageable and inspectable procedure
candidates that can be composed, reused, and refined by later evidence.  This
supports the broader claim that structural correspondence can guide learning
from worked transformations rather than merely reproduce isolated target values.

AST distance gives the model a concrete measure of how structurally close two
parts of an expression are.  Our experiments show that this ordering improves
computational search, but they do not establish that people search in exactly
the same way.  The model therefore predicts that people should prefer
explanations using values and operations from corresponding, nearby regions of
an expression and consider distant or unstated knowledge later.  A behavioral
study could vary where the relevant evidence appears while keeping the intended
arithmetic relationship fixed, then compare participants' explanations, errors,
and response times with the model's predictions.

\subsection{Broader Architectural Context}
\label{sec:broader}

The evaluated mechanism explains how observed values were produced and turns
those explanations into executable candidate procedures.  DIPL provides a
broader computational theory in which it could support multi-step learning from
multiple examples \citep{weitekamp2025dipl}.  Where-learning would identify and
bind a procedure's inputs, while when-learning would acquire the conditions for
applying it.  Together, these processes would remove the supplied-role
assumption in our transfer test and select procedures within new problem states.

This division of labor also gives later examples a principled role.  A single
transformation may support several successful explanations even after guidance
has sharply reduced search.  The present mechanism supplies ordered,
inspectable candidates; a broader learner can retain those consistent across
examples and reject first-case coincidences.  Our transfer analysis illustrates
this interaction: one new instance selected a candidate that applied to the
next instance in all 16 families.  Integrating where- and when-learning would
extend this result to contextual selection and multi-step problem solving,
allowing a simulated learner to accumulate and generalize procedural knowledge.

Practice could also change what the learner searches over.  The current ASTs
already represent nested expressions, and our experiments compose local
explanations into complete procedures and apply them to new values.  A further
step would store a learned procedure as a reusable building block within a
longer solution.  For example, once a learner has acquired a procedure for
combining like terms, it could retrieve that procedure within a larger
expression rather than explain the coefficient calculation again.  Where- and
when-learning would identify its arguments and the contexts in which reuse is
appropriate.  This could reduce repeated search as experience accumulates;
learning such a hierarchy of reusable procedures was not evaluated here.
Testing this extension on less structured problems, where the relevant
transformation must itself be identified, would connect the present
worked-step setting to broader mathematical problem solving.

\section{Conclusion}

Learning from a worked transformation requires explaining its changes, not
merely reproducing its result.  The proposed mechanism combines structure
mapping, relational locality, and grounded abduction to prioritize explanations
supported by the demonstrated expression.  It recovered all 120 intended local
procedures, ranked 104 first rather than an unguided median of 32, and examined
93.8\% fewer value--operation combinations.  Against randomized visible-first
search, it ranked 104 rather than a median of 62 first and examined 63.6\% fewer
combinations.  Thus correspondence-relative order contributes beyond merely
preferring displayed knowledge.  Value-distance guidance matched full guidance's
intended-first count at lower cost on the original corpus.  Operation
distance added a benefit when visible operators competed, reducing search by
43.5\% relative to value-distance guidance in that analysis.  The contribution
is therefore a structural account of relevance whose benefits depend on the
evidence available in the expression.

These priorities survived composition: the intended complete procedure ranked
first in 56 of 70 transformations rather than an unguided median of 15.5, and
preceding alternatives fell from 81,116 to 133.  Top-ranked candidates produced
28 of 32 new targets, and one additional example selected a candidate that
transferred in all 16 families.  These results yield a testable cognitive
prediction rather than an established account of human behavior: people's
explanations should favor corresponding and local evidence.  Taken together,
the findings show that structural correspondence can make abductive procedure
learning from worked examples both computationally tractable and interpretable
without placing the demonstrated procedure out of reach.

{\parindent -10pt\leftskip 10pt\noindent
\bibliographystyle{cogsysapa}
\bibliography{references}
}

\end{document}